\documentclass[onecolumn]{nestpaper}

\usepackage{empheq}
\usepackage{xargs}

\title{A Micro-Macro Model of Encounter-Driven Information Diffusion in Robot Swarms}

\author{%
  Davis~S.~Catherman\orcidID{0000-0001-9161-0223}\\
  Robotics Engineering, Worcester Polytechnic Institute\\
  Worcester, MA, USA\\
  \email{dscatherman@wpi.edu}
  \and
  Carlo~Pinciroli\orcidID{0000-0002-2155-0445}\\
  Robotics Engineering, Worcester Polytechnic Institute\\
  Worcester, MA, USA\\
  \email{cpinciroli@wpi.edu}}

\begin{document}

\published{The 15th International Conference on Swarm Intelligence (ANTS 2026)}

\maketitle

\blfootnote{DISTRIBUTION STATEMENT A. Approved for public release; distribution is unlimited. OPSEC\#10389}

\begin{abstract}
  In this paper, we propose the problem of \emph{Encounter-Driven Information Diffusion (EDID)}. In EDID, robots are allowed to exchange information only upon meeting. Crucially, EDID assumes that the robots \emph{are not allowed to schedule their meetings}. As such, the robots have no means to anticipate when, where, and who they will meet. As a step towards the design of storage and routing algorithms for EDID, in this paper we propose a model of information diffusion that captures the essential dynamics of EDID. The model is derived from first principles and is composed of two levels: a \emph{micro} model, based on a generalization of the concept of `mean free path'; and a \emph{macro} model, which captures the global dynamics of information diffusion. We validate the model through extensive robot simulations, in which we consider swarm size, communication range, environment size, and different random motion regimes. We conclude the paper with a discussion of the implications of this model on the algorithms that best support information diffusion according to the parameters of interest.
\end{abstract}

\section{Introduction}
\label{sec:Introduction}

Effective communication is an essential mechanism to facilitate multi-robot coordination and cooperation~\citep{Gielis2022}. Multi-robot applications in which communication plays a pivotal role are diverse and include warehouse logistics~\citep{Hnig2019PersistentAR}, agriculture~\citep{albiero2021swarm}, climate monitoring~\citep{hafid2024analyzing}, firefighting~\citep{penders2011robot}, space exploration~\citep{Vivek2025RA}, and underground mining~\citep{tan2024evaluating}. Extensive work exists on the constraints communication imposes on robot swarms, including maintaining connectivity~\citep{Garzon2016AMS}, dealing with bandwidth~\citep{nunnally2012human} and power~\citep{Zhang2018PerformanceBO} limitations, and designing robust countermeasures to message loss~\citep{Manfredi2020Trans}.

This paper focuses on \emph{Encounter-Driven Information Diffusion (EDID)}, a peculiar form of communication defined by two distinctive conditions. First, the robots are assumed to have a short communication range relative to the size of the environment. Thus, they can exchange messages only upon meeting each other. Second, robot motion is dictated by factors that act \emph{independently} from communication needs. Consequently, robots cannot necessarily predict who and when they will meet in the future. The latter condition sets EDID apart from prior work on sparse swarms \citep{Tarapore2020SparseRS} and opportunistic communication \citep{Cheraghi2020,Mox2024,Cladera2024}, in which the robots maintain agency in their motion patterns.

Compared to other types of constrained communication, research on EDID has received little attention. Yet, EDID appears in many robotics engineering applications (e.g., underwater operations~\citep{Champion2015}, planetary exploration~\citep{delaCroix2024MultiAgentAF}), and in the study of natural collective behavior through wearable devices~\citep{Halgas2023,Olguin2007,Olguin2009}.

Ultimately, the goal of this work is to produce a model for information diffusion that will inform the design of a new class of algorithms for information management that target EDID. EDID is different from typical scenarios that involve message loss, bandwidth limitations, and sparsity. At its most fundamental level, the fact that robot motion is not a controllable variable means that information diffusion cannot rely on algorithms that optimize network topology~\citep{Kantaros2016,Luo2019,Majcherczyk2018,Vandermeulen2018} or adapt information routing~\citep{Cladera2024,Mox2024}. Stochastic aspects dominate the dynamics, such as when, how long, and which robots meet during a mission. Local encounter rates do not map linearly to global information diffusion, making it necessary to develop models that combine microscopic (robot-level) and macroscopic (swarm-level) factors.

In this paper, we propose the first model of EDID. Our model links the microscopic aspects of individual robot motion with the macroscopic dynamics of information diffusion. At the microscopic level, we identify the concept of \emph{mean free path}~\citep{Feynman2011} as an informative model component. At the macroscopic level, our model reveals that the dynamics of information diffusion presents a bifurcation driven by parameters such as robot density and motion patterns. We derive the model from first principles and validate it in an extensive campaign of physics-accurate simulated experiments, exploring the role of swarm size, communication range, environment size, and various types of random motion.

\section{The Microscopic Model}
\label{sec:microscopic-model}

Our model is composed of two parts: a \emph{microscopic} model that captures the dynamics of individual encounters, and a \emph{macroscopic} one that characterizes aggregate behavior.

\paragraph{Encounters and mean free path}
Consider a swarm of $N$ robots capable of communication within a range $C$. The robots are distributed in a square area of side $L$ and move according to some form of random walk (we will specify this aspect in Sec.~\ref{sec:exper-eval}). An \emph{encounter} occurs when two or more robots are within communication range, making it possible for messages to be exchanged.

Our microscopic model originates from the concept of \emph{mean free path}~\citep{Feynman2011}, i.e., the distance a robot covers between two encounters on average. The mean free path models the microscopic dynamics of colliding gas particles. The use of mean-free-path and kinetic-gas-theory–inspired reasoning to model inter-robot encounters is not new in swarm robotics. For example, \citet{wahby2019collective} employ encounter statistics derived from this physical intuition to adapt aggregation behavior to dynamic swarm densities and environmental conditions.

\begin{figure*}[t]
    \centering
    \includegraphics[height=5.0cm]{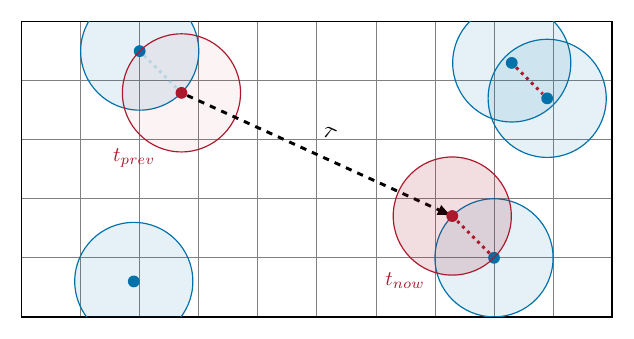}
    \caption{A representation of the mean free path $\tau$ as the average time between subsequent interactions between an agent (depicted in red) and surrounding agents (in blue).}
    \label{fig:tau}
\end{figure*}

In our scenario, encounters occur when two robots cross each others' communication range; notably, an encounter does not necessarily imply a change in direction, i.e., a physical collision. Thus, our microscopic model will include a corrective term. A diagram depicting the mean free path is reported in Fig. \ref{fig:tau}.

\paragraph{Deriving the ideal mean free path}
We first derive the expression of the `ideal' mean free path, $l_\mathrm{ideal}$, which corresponds to the definition found in gas particle modeling. To derive $l_\mathrm{ideal}$, we resort to a probabilistic argument:
\begin{itemize}
\item The probability that a robot encounters another along the mean free path is $P(\text{encounter}) = dx /l_\mathrm{ideal}$ where $dx < l_\mathrm{ideal}$ is the distance covered by the robot in a time $dt$.
\item  Another approach to calculate $P(\text{encounter})$ involves considering when two robots intersect each others' communication range. The average number of robots that can be encountered per unit of area is $(N-1)/L^2$; along a path of length $dx$, we expect to encounter $(N-1) / L^2 \cdot dx$ robots. As the robot moves in the 2D environment, it communicates with all the robots within a range $C$, so $P(\text{encounter}) = C \cdot (N-1) \cdot dx / L^2$.
\end{itemize}
Putting the two expressions together yields:
\begin{equation*}
  P(\text{encounter}) =
  \frac{dx}{l_\mathrm{ideal}} =
  \frac{C \cdot (N-1)}{L^2} dx
  \Rightarrow
  l_\mathrm{ideal} = \frac{L^2}{C \cdot (N-1)}.
\end{equation*}
Because our macroscopic model will capture the time dynamics of information diffusion, we will need the microscopic model to be time-based. To achieve this, we consider the mean time between two encounters, denoted by $\tau_\mathrm{ideal}$. If the robots move at a constant speed $V$, then we simply have
\begin{equation}
  \label{eq:tauideal}
  \tau_\mathrm{ideal} = \frac{l_\mathrm{ideal}}{V} = \frac{L^2}{C \cdot (N-1) \cdot V}
\end{equation}

The effect of the parameters of interest $\langle L,N,C,V \rangle$ are not surprising. As the environment side $L$ increases, the mean time between interactions increases super-linearly. Conversely, as the swarm size $N$ increases, $\tau$ drops. Analogously, a larger communication range $C$ facilitates more frequent encounters. These effects are related to robot density --- the higher the density, the shorter the times between robot encounters. The role of the speed $V$ is to promote mixing: Higher speeds result in lower $\tau_\mathrm{ideal}$.

\paragraph{The generalized mean free path}
While $\tau_{ideal}$ is derived from the kinetic theory of gases, it assumes point-mass particles and instantaneous collisions. Real robotic encounters differ in two critical ways: they have a finite duration (determined by $C/V$) during which the robots remain in range, and they occur within a bounded environment where boundary effects influence density. To bridge the gap between the idealized gas model and physical reality, we treat Eq. \eqref{eq:tauideal} as a semi-empirical formulation. We introduce a factor $\beta = \tau / \tau_\mathrm{ideal}$ as a phenomenological parameter derived via regression from the experimental data in Fig. \ref{fig:micro_validation}. This additional parameter serves as a corrective term that accounts for the non-zero duration of communication and the geometric constraints of the arena. In our macroscopic model, we will use $\tau = \beta \cdot \tau_\mathrm{ideal}$.

\section{The Macroscopic Model}
\label{sec:macroscopic-model}

\begin{figure*}[t]
    \centering
    {\includegraphics[height=4.0cm,width=4.0cm]{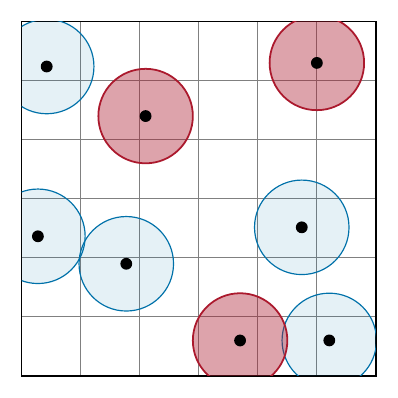}}
    {\includegraphics[height=4.0cm,width=4.0cm]{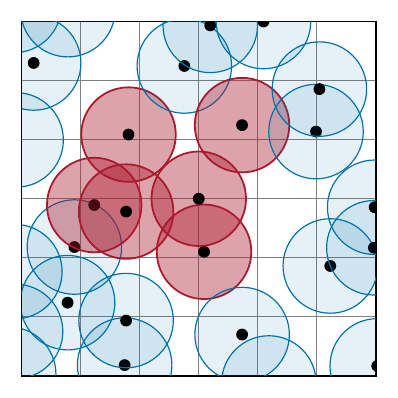}}

    \caption{Our macro model identifies two possible regimes: low-density, well mixed (left) and high-density, not-mixed (right).}
    \label{fig:macro_model_low_high_density_mixture}
\end{figure*}

We now derive the macroscopic model from first principles. Let us call $P_n(t)$ the probability that exactly $n$ robots are informed at time $t$, $\sum_{n=0}^N P_n(t) = 1$. The evolution of $P_n(t)$ is determined by robot encounters. We call $W_{n \to n+1}$ the transition rate (probability per unit time) that the swarm jumps from $n$ to $n+1$ informed robots. Given these quantities, we can write the master equation
\begin{equation*}
  \frac{dP_n}{dt} =
  P(\text{gain}) - P(\text{loss}) =
  W_{n-1 \to n} P_{n-1} - W_{n \to n+1} P_n.
\end{equation*}
In the above equation, we consider each state with $n$ informed robots as a `bin', with probabilities that flow from one bin to another (i.e., gain and loss) as robots become informed. The master equation allows us to calculate the dynamics of the expected number of informed robots $\langle n \rangle = \sum_{n=0}^N n P_n$:
\begin{equation}
  \label{eq:midstep}
  \frac{d\langle n \rangle}{dt}
  = \sum_{n=0}^N n \frac{dP_n}{dt}
  = \underbrace{\sum_{n=0}^N n W_{n-1 \to n} P_{n-1}}_{(*)} - \sum_{n=0}^N n W_{n \to n+1} P_n.
\end{equation}
We modify term $(*)$ by introducing the change of variable $m = n-1$:
\begin{align*}
  \sum_{m=-1}^{N-1}(m+1)W_{m \to m+1}P_m = \sum_{m=0}^{N}(m+1)W_{m \to m+1}P_m = \sum_{n=0}^N (n+1) W_{n \to n+1} P_n,
\end{align*}
where
\begin{inparaenum}
\item the second step exploits the facts that $P_{m} = 0$ for $m=-1$ and that $W_{N \to N+1} = 0$
\item the third step simply relabels $m$ as $n$
\end{inparaenum}
Plugging the resulting expression into Eq. \eqref{eq:midstep} yields
\begin{equation}
  \label{eq:master}
  \frac{d\langle n \rangle}{dt} = \sum_{n=0}^N (n+1)W_{n \to n+1} P_n - \sum_{n=0}^N n W_{n \to n+1} P_n = \sum_{n=0}^N W_{n \to n+1} P_n
\end{equation}
This equation depends on the specific form of $W_{n \to n+1}$. Intuitively, we can identify two potential dynamics acting as `extremes' (see Fig. \ref{fig:macro_model_low_high_density_mixture})
\begin{inparaenum}
\item A low-density, well-mixed regime
\item A high-density, not-mixed regime
\end{inparaenum}
We hypothesize that most real deployments will display dynamics that blend elements of these extremes. Next, we study these extreme scenarios individually. In both cases, we consider the rate of change of the fraction of informed robots $I = \langle n \rangle / N$ over time, i.e., $dI / dt = d \langle n \rangle / dt$. 

\paragraph{Low-density, well-mixed regime}
\label{sec:low-density-well-mixed}
In a low-density regime, the robots rarely meet each other. Thus, every meeting is important to diffuse information across the swarm. If the robots are also well-mixed, when a meeting occurs, it is likely to involve two robots who rarely (or never) met before. This phenomenon can be captured by considering a meeting between an informed robot and an uninformed one. The informed robot meets other robots at a rate $1/\tau$; due to well-mixedness, the uninformed robot can be chosen uniformly at random, i.e., with a probability $(N-n) / N$. Thus, the rate of change of informed robots follows the law:
\begin{equation*}
  \frac{d\langle n \rangle}{dt} =
  \sum_{n = 0}^N \underbrace{n \cdot \frac{1}{\tau} \cdot \frac{N - n}{N}}_{W_{n \to n+1}} \cdot P_n =
  \frac{1}{\tau} \left[ \langle n \rangle - \frac{\langle n^2 \rangle}{N} \right].
\end{equation*}
In the mean field limit, i.e., for $N \to \infty$, $\mathrm{Var}(n) = \langle n^2 \rangle - \langle n \rangle^2 \sim O(\sqrt{N})$---fluctuations become negligible---and $\langle n^2 \rangle \sim \langle n \rangle^2$. We can approximate
\begin{equation*}
  \frac{d\langle n \rangle}{dt} \approx
  \frac{1}{\tau} \left[ \langle n \rangle - \frac{\langle n \rangle^2}{N} \right] =
  \frac{\langle n \rangle}{\tau} \left[1  - \frac{\langle n \rangle}{N} \right]
\end{equation*}
which, using $I = \langle n \rangle / N$, becomes the well-known logistic law \citep{feller1940logistic}:
\begin{equation}
  \label{eq:logistic}
  \frac{dI}{dt} = \frac{I(1-I)}{\tau}.
\end{equation}

\paragraph{High-density, not-mixed regime}
\label{sec:high-density-not-mixed}
In a high-density, not-mixed regime, information propagates in waves and is conveyed redundantly by several robots. An uninformed robot is likely to receive several copies of the same message, making the average amount of information exchanged per-meeting much lower than in the logistic case. In this regime, the limiting factor is not the number of uninformed robots, as in the logistic case, which produced the term $(N-n)/N$. Rather, it is how \emph{surprising} a message is. To model this phenomenon, we need a factor $\eta(n/N)$ to express how surprising a message is for a robot. The new factor must satisfy two basic constraints
\begin{inparaenum}
\item $\eta(n/N)$ is continuous and monotonically decreasing
\item $\eta(n/N) \to 0$ as $n/N \to 1$
\end{inparaenum}
There are infinite possible choices for $\eta(n/N)$. Drawing inspiration from information theory, we propose the ansatz $\eta = - \ln (n/N)$, which expresses the Shannon self-information \citep{zheng2022shannon} of having $n$ informed robots. Self-information indeed quantifies how surprising an event is. When $n/N$ is small, receiving a message from an informed robot is surprising (high $\eta$); when $n/N \to 1$, it is completely predictable ($\eta \to 0$). Now we can derive our model for this regime:
\begin{equation*}
  \frac{d\langle n \rangle}{dt} = \sum_{n = 0}^N \underbrace{n \cdot \frac{1}{\tau} \cdot \left(-\ln \frac{n}{N} \right)}_{W_{n \to n+1}} \cdot P_n.
\end{equation*}
Using $I = \langle n \rangle /N$, the final form of this equation is a Gompertz law \citep{winsor1932gompertz}:
\begin{equation}
  \label{eq:gompertz}
  \frac{dI}{dt} = \frac{1}{\tau} I \ln \frac{1}{I}.
\end{equation}

\paragraph{Combined model}
\label{sec:combined-model}
The next step is linearly combining Eq.~\eqref{eq:logistic} and Eq.~\eqref{eq:gompertz} through a weight $\lambda \in [0,1]$ that governs the mixture of the two regimes:
\begin{equation}
  \label{eq:combined}
  \frac{dI}{dt} = \lambda\frac{I}{\tau}\ln \frac{1}{I} + \left( 1 - \lambda \right) \frac{I(1-I)}{\tau}.
\end{equation}
This combined model cannot, in general, be integrated into a closed-form solution. The only two closed-form solutions exist for $\lambda = 0$ (pure logistic) and $\lambda = 1$ (pure Gompertz):
\begin{align}
  I(t)_{\lambda = 0} &= \frac{I_0}{I_0 + (1 - I_0) \cdot e^{-(t-t_0) / \tau}} \label{eq:ilogistic}\tag{LGSTC} \\
  I(t)_{\lambda = 1} &= \exp\left(-\ln I_0 \cdot e^{-(t-t_0) / \tau}\right) \label{eq:igompertz}\tag{GMPRZ}
\end{align}
where $I_0 = 1/N$ is the number of informed robots at the start of an experiment. Because Eq. \eqref{eq:combined} is a nonlinear differential equation, its solution is not, in general, the weighted sum of the individual logistic and Gompertz solutions. Nevertheless, both limiting curves are smooth sigmoids that differ only slightly in shape under the parameter ranges we observed, making a linear-combination approximation practically accurate for our data (the typical root-mean-square error is $\approx 10^{-2}$). For this reason, it is practically acceptable to approximate the solution of Eq. \eqref{eq:combined} as a linear combination of \eqref{eq:ilogistic} and \eqref{eq:igompertz}. This is our final model:
\begin{empheq}[box=\fbox]{align}
    I(t) \approx &\; \lambda \exp\left(-\ln I_0 \cdot e^{-(t-t_0) / \tau}\right) + \label{eq:final}\tag{MDL}\\
    &\; + (1-\lambda) \frac{I_0}{I_0 + (1 - I_0) \cdot e^{-(t-t_0) / \tau}}. \notag
\end{empheq}

\begin{table}[t]
  \centering
  \caption{
    For CRW, LW, and Hybrid walks, the parameters were studied across a range specified by $(start, stop, step)$.
  }
  \label{tab:experiment_params}
  \begin{tabular}{p{2cm}p{2.5cm}p{7cm}}
    \toprule
    \textbf{Parameter} & \textbf{Default} & \textbf{Parameter Study}\\
    \midrule
    $N$       & 20   & \{ 6, 10, 14, 18, 22, 26, 30, 34, 38, 42, 46, 50 \} \\
    \midrule    
    $C$ (m)   & 10   & \{ 6, 10, 14, 18, 22, 26, 30, 34, 38, 42, 46, 50 \} \\
    \midrule
    $L$ (m)   & 200  & \{ 40, 60, 80, 100, 120, 140, 160, 180, 200, 220 \} \\
    \midrule
    $V$ (m/s) & 0.05 & \{ 0.05 \}\\
    \midrule
    Walk      & CRW ($\rho=0.7$) & 

    $\begin{aligned}[t] 
      \{\hspace{1mm}CRW&(\rho=(0.1, 0.9, 0.1)),\\
      LW&(\alpha=(1.4, 2.8, 0.1)),\\
      Hybrid&(\rho=(0.2, 0.8, 0.2),  \alpha=(1.4, 2.4, 0.2))\hspace{1mm}\} \\
    \end{aligned}$\\
    \bottomrule
  \end{tabular}
\end{table}

\section{Experimental Evaluation}
\label{sec:exper-eval}

To validate our model, we ran an extensive set of simulated experiments in the ARGoS multi-robot simulator~\citep{PinciroliSI2012}. It is worth highlighting that our simulations are physically accurate with respect to real-world dynamics, as opposed to simple numerical simulations of mass-less particles. We opted for this approach to fully evaluate the correspondence between our model and realistic mobility dynamics.

\paragraph{Experimental Setup}
The parameters we analyzed are summarized in Tab.~\ref{tab:experiment_params}. For each configuration, we adjusted a single parameter while keeping the other parameters at the reported default value. We ran 40 repetitions for each parameter configuration, for a total of 3,160 simulations. The duration of each simulation was 100 simulated hours with a step resolution of 1 second. At the start of each simulation, the robots are uniformly distributed in an empty arena of size $L$. Any pair of robots experience an encounter with each other while the separation distance is less than the communication range. To study information diffusion during a simulation, we examine a single robot that generates a new message every hour for the first 50 hours and record the duration for each peer to receive the message. We utilize the standard definitions of Correlated Random Walks (CRW) and Levy Walks (LW) as described in \citet{Dimidov2016RandomWI}. We vary the direction correlation parameter $\rho \in [0, 1]$ for CRW and the step length distribution exponent $\alpha \in [1, 2]$ for LW. As $\rho\rightarrow0$, the motion is increasingly brownian, meanwhile the LW step distribution is characterized by long relocations as $\alpha\rightarrow0$. The hybrid walk then combines the two components in one random motion. 

\begin{figure*}[tb]
  \centering
  \includegraphics[width=1.0\linewidth]{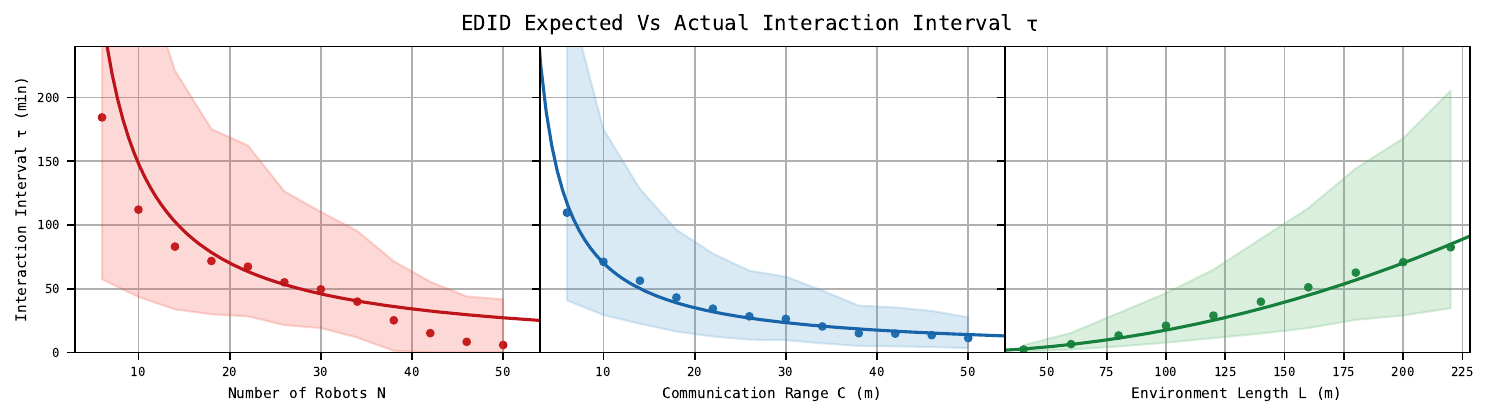}%
  \caption{Interaction interval for swarm configurations considering the number of robots ($N$), communication range ($C$) and the square-environment length ($L$) given the robot velocity ($v$). The plot reports means and inter-quartile ranges.}
  \label{fig:micro_validation}%
\end{figure*}

\paragraph{Microscopic Model}
We performed experiments to validate our microscopic model ($\tau$) across the parameters of interest listed in Tab.~\ref{tab:experiment_params}. The results are plotted in Fig.~\ref{fig:micro_validation}. The model shows a good match between predicted values and experimentally derived ones. For brevity, we omitted the plots in which we varied the random walks, because we observed negligible effects.

\begin{figure}[t]
  \centering
  \subfloat[The message propagation over time for multiple experimental configurations with varying parameters $N$ (red), $C$ (blue), and $L$ (green). The dashed line represents the output of the fitted blended model for each curve.
    \label{fig:4_propagation_fit_params}]%
  {%
    \includegraphics[height=5.6cm]{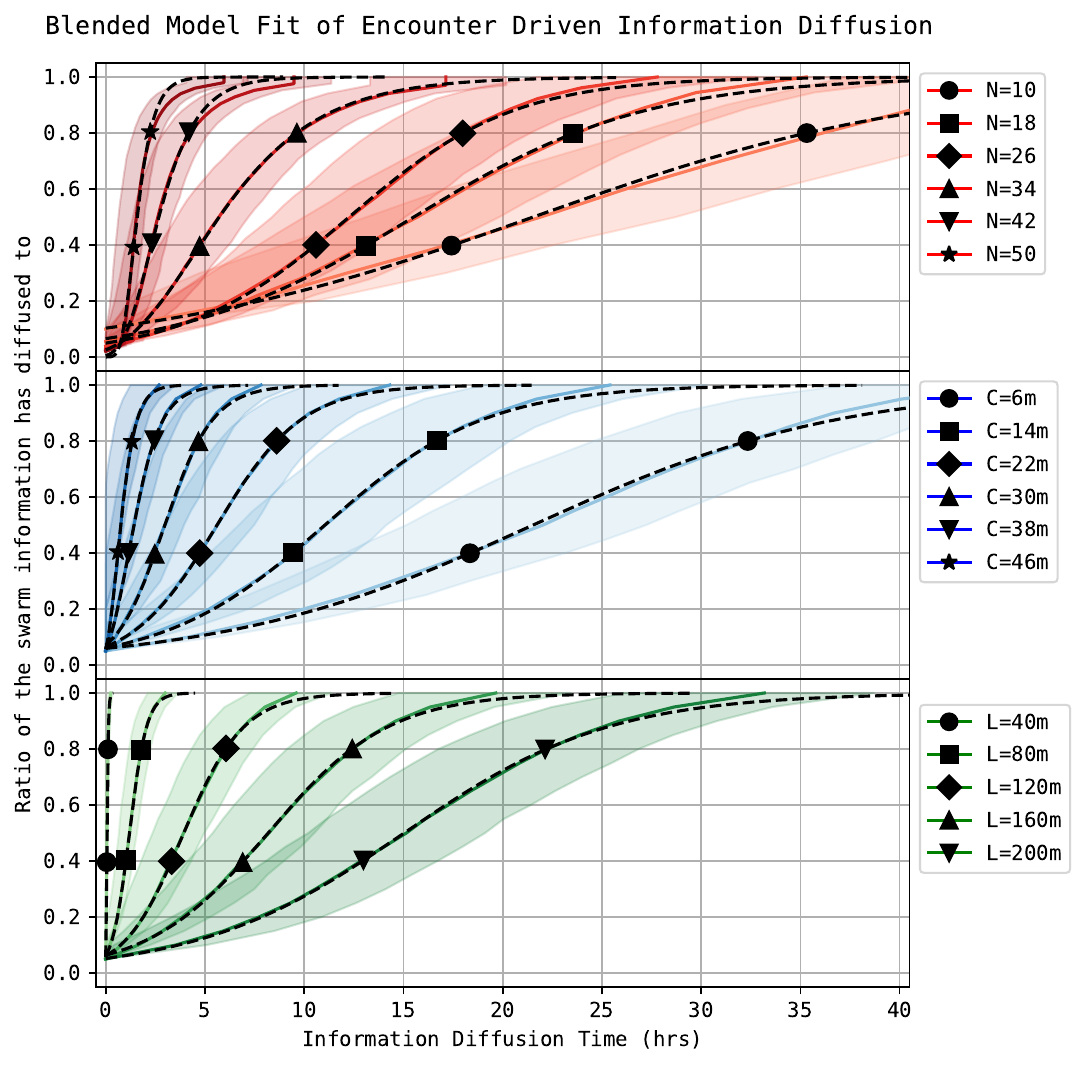}%
  }
  \hfill
  \subfloat[The message propagation over time for multiple experimental configurations with varying random walk parameters: $CRW(\rho)$ (red), $LW(\alpha)$ (blue), and $Hybrid(\rho,\alpha)$ (green). The dashed line represents the output of the fitted blended model for each curve.
    \label{fig:4_propagation_fit_walks}]%
  {%
    \includegraphics[height=5.6cm]{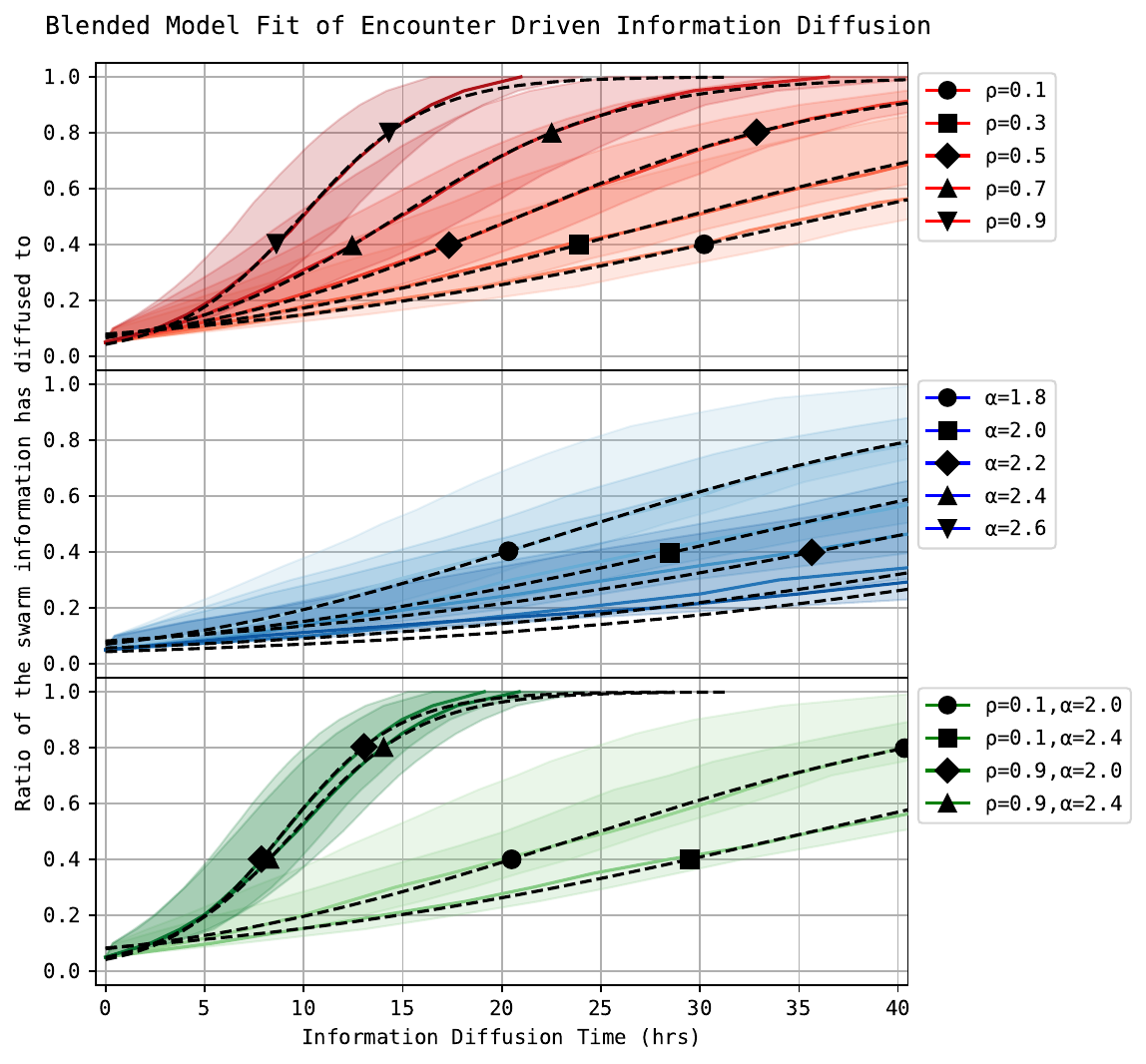}%
  }
  \caption{Message propagation over time for a subset of the various experimental conditions.}
\end{figure}

\begin{figure*}[t]
    \centering

    {\includegraphics[width=\linewidth]{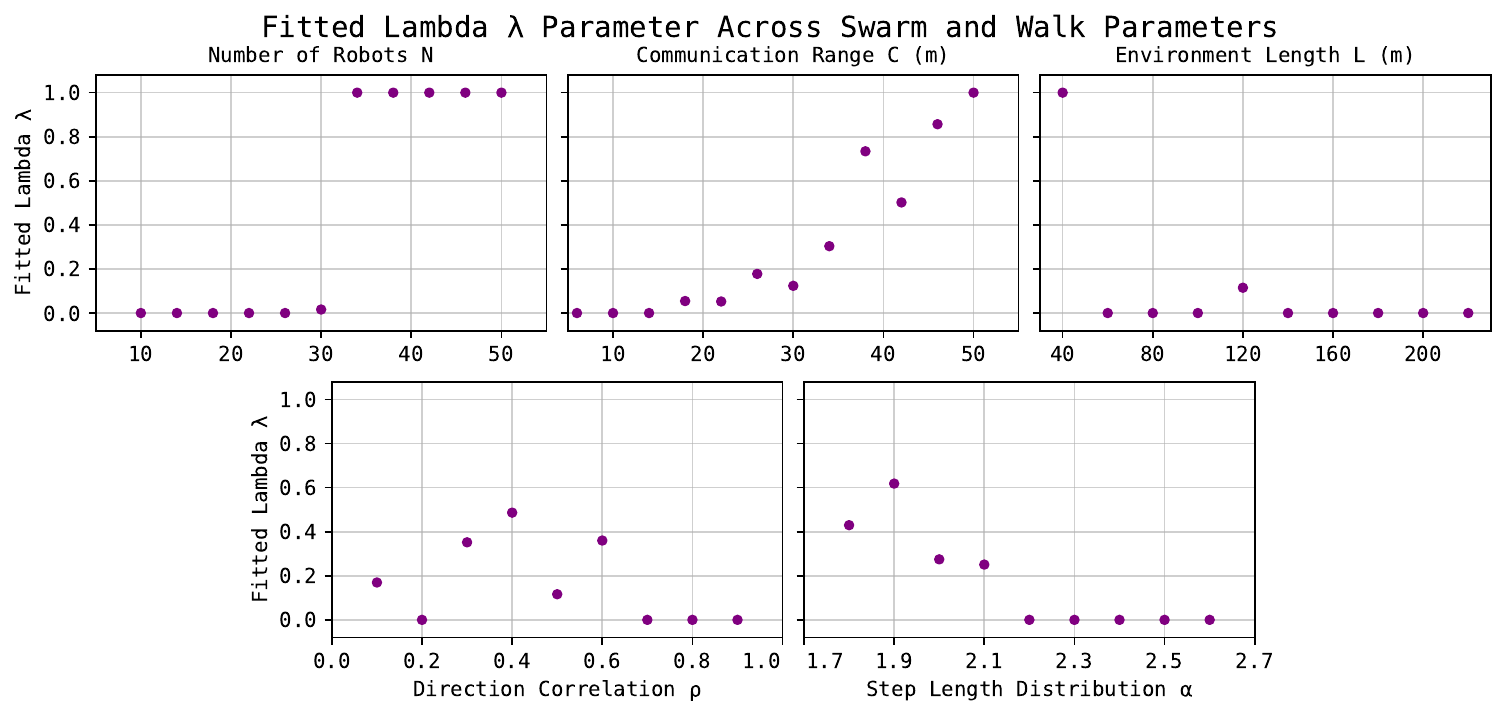}}
    
    \caption{Parameter $\lambda$ for blending the Logistic and Gompertz as fit across all IVs.}
    \label{fig:logistic_gompertz_alpha_blend}
\end{figure*}

\paragraph{Macroscopic Model: Parameter Fit}
We fit the parameters of the macroscopic model to the data we collected. The results show a remarkable accuracy across the experiment conditions. We present the fits across the swarm and environment parameters in Fig.~\ref{fig:4_propagation_fit_params} and across the random walks in Fig.~\ref{fig:4_propagation_fit_walks}. These plots demonstrate that our model captures well a wide span of experimental conditions. The blending parameter $\lambda$ offers a physical interpretation of the swarm's `informational viscosity.' As shown in Fig. \ref{fig:logistic_gompertz_alpha_blend}, $\lambda$ acts as a sliding scale between the well-mixed regime (Logistic, $\lambda \approx 0$) and the redundant, wave-propagation regime (Gompertz, $\lambda \approx 1$). We observe that $\lambda$ increases sharply with communication range $C$. Physically, a larger communication radius increases local network density, causing robots to receive redundant transmissions from neighbors they have already met; this saturation of local information mirrors the `surprisal' decay modeled by the Gompertz function. Conversely, as the environment size $L$ increases or the motion becomes more directed (higher $\rho$), $\lambda$ trends toward 0. In these high-mobility or sparse scenarios, the swarm approaches the `well-mixed' ideal: encounters are rare events between agents who are unlikely to share immediate neighbors, making the Logistic model the dominant driver of diffusion.

\paragraph{Macroscopic Model: Propagation Time}
We studied the effect of the parameters of interest on the propagation time of the messages. The results for $\langle N,C,L \rangle$ are reported in Fig.~\ref{fig:4_propagation_fit_params}. Not surprisingly, as the density of the robots increases, the propagation time decreases. As for random walks, the results are reported in Fig. \ref{fig:4_propagation_fit_walks}. The propagation time varied notably across different random walks. LWs provided long propagation times with multiple messages not having fully propagated when the experiment ended. For increasing $\rho$ there is a noticeable decrease in wall time. Additionally, for decreasing $\alpha$, which is related to having a heavier tail for step length, there appears to be decreasing wall time. For both walks, a generally more straight path movement related to lower propagation times. These latter results are obtained with $\rho=0.7$, helping to account for likely path deviations or obstacle avoidance. This reaffirms that motion dynamics, and not just communication density, is relevant to understanding the lifespan of a message.

\section{Conclusion}
\label{sec:Conclusion}

In this paper, we proposed a novel micro-macro model for Encounter-Driven Information Diffusion (EDID) in robot swarms. The model is composed of two parts: a microscopic model, based on the concept of \emph{mean free path}, that estimates the expected rate of encounters among robots. The macroscopic model, on the other hand, targets swarm-level information diffusion. The main insight in our macroscopic model is the identification of two regimes that dominate the diffusion dynamics as a consequence of robot density. For high communication density, our model follows the Gompertz curve, which captures the diminishing returns of information transfer due to message replication. For low communication density, the logistic model proves accurate. Our final model blends these two extremes and provides excellent fit to the experimental data we collected.

This model, while simple, sheds light on the design of algorithms for information diffusion in robot swarms. Specifically, if the goal is to achieve fast and uniform diffusion, uncorrelated random walks produce Gompertz-like dynamics. If, conversely, the goal is to diffuse in a hierarchical, controllable manner, correlated motion provides better results due to the logistic dynamics it produces.

Our model contributes to the broader literature on population dynamics in swarm robotics and ecology by establishing a novel connection between density-driven regimes and functional forms of macroscopic diffusion. While prior work has extensively applied logistic dynamics to robot swarms under well-mixed assumptions~\citep{hamann2014derivation,valentini2016collective,reina2015design}, and the Gompertz equation has been used to model tumor growth~\citep{laird1964dynamics,benzekry2020population} and epidemic spreading on networks~\citep{estrada2022gompertz}, no existing work applies Gompertz dynamics to robot swarm information diffusion or models a continuous bifurcation between these regimes. The blending parameter $\lambda$ in our model addresses this gap by capturing how communication density drives a transition from well-mixed logistic spreading to redundant Gompertz-like propagation~\citep{tjorve2017gompertz}. Our micro-macro derivation extends kinetic-theory-based encounter models~\citep{wilson2020closed,kernbach2011diffusion} beyond timing predictions to reveal the specific functional form of growth curves, connecting mean free path not merely to diffusion rates but to the qualitative character (logistic versus Gompertz) of the macroscopic dynamics~\citep{hamann2008framework,elamvazhuthi2020mean}.

In the context of EDID, our model provides a way to characterize the expected diffusion behavior given measurable parameters, such as the size of the environment, the number of robots involved, and the communication range. We plan to use our model as a component in a new class of storage-and-routing algorithms for EDID scenarios.

\section*{Acknowledgments}
This work was supported by the Automotive Research Center (ARC), a US Army Center of Excellence for modeling and simulation of ground vehicles, under Cooperative Agreement W56HZV-24-2-0001 with the US Army DEVCOM Ground Vehicle Systems Center (GVSC).

\bibliographystyle{plainnat}
\bibliography{refs}

\end{document}